\def\year{2018}\relax
\documentclass[letterpaper]{article} 
\usepackage{aaai18}  
\usepackage{times}  
\usepackage{helvet}  
\usepackage{courier}  
\usepackage{url}  
\usepackage{graphicx}  
\usepackage{url}
\usepackage{multirow}
\usepackage{epsfig}

\usepackage{subfig}
\usepackage{listings}
\usepackage{amsmath,amssymb,amsbsy,bm,amsthm}

\theoremstyle{definition}

\DeclareMathOperator*{\argmax}{arg\,max}
\DeclareMathOperator*{\argmin}{arg\,min}

\begin{document}
%
\title{Explicit Reasoning over End-to-End Neural Architectures \\for Visual Question Answering }

\author{Somak Aditya, Yezhou Yang and Chitta Baral\\
School of Computing, Informatics and Decision Systems Engineering\\
Arizona State University\\
\{saditya1,yz.yang,chitta\}@asu.edu}

\maketitle
\begin{abstract}
Many vision and language tasks require commonsense reasoning beyond data-driven  
image and natural language processing. Here we adopt Visual Question Answering (VQA) as an example task, where a system is expected to answer a question in natural language about an image.  Current state-of-the-art systems attempted to solve the task using deep neural architectures and achieved promising performance. However, the resulting systems are generally opaque and they
struggle in understanding questions for which extra knowledge is required. In this paper, we present an explicit reasoning layer on top of a set of penultimate neural network based systems. The reasoning layer enables  reasoning and answering questions where additional knowledge is required, and at the same time provides an interpretable interface to the end users. 
Specifically, the reasoning layer adopts a Probabilistic Soft Logic (PSL) based engine to reason over a basket of inputs: visual relations, the semantic parse of the question, and background ontological knowledge from word2vec and ConceptNet. Experimental analysis of the answers and the key evidential predicates generated on the VQA dataset validate our approach.

\end{abstract}

\section{Introduction}



Many vision and language tasks are considered as compelling ``AI-complete'' tasks which require multi-modal knowledge beyond a single sub-domain. 
One such recently proposed popular task is Visual Question Answering (VQA) by \cite{antol2015vqa}, which requires a system to generate natural language answers to free-form, open-ended, natural language
questions about an image.  Needless to say, this task is extremely challenging since it falls on the junction of three domains in Artificial Intelligence: image understanding, natural language understanding, and commonsense reasoning. With the rapid development in deep neural architectures for image understanding, end-to-end networks  trained from pixel level signals together with word embeddings of the posed questions to the target answer, have achieved promising performance \cite{malinowski2015ask,gao2015you,lu2016hierarchical}. Though the resulting answers are impressive, the capabilities of these systems are still far from being satisfactory.  We believe the primary reason is that many of these systems overlook the critical roles of natural language understanding and commonsense reasoning, and thus fail to answer correctly when additional knowledge is required.  

\begin{figure*}[!htb]
     \centering
     \subfloat{\includegraphics[width=0.75\textwidth,height=0.17\textheight]{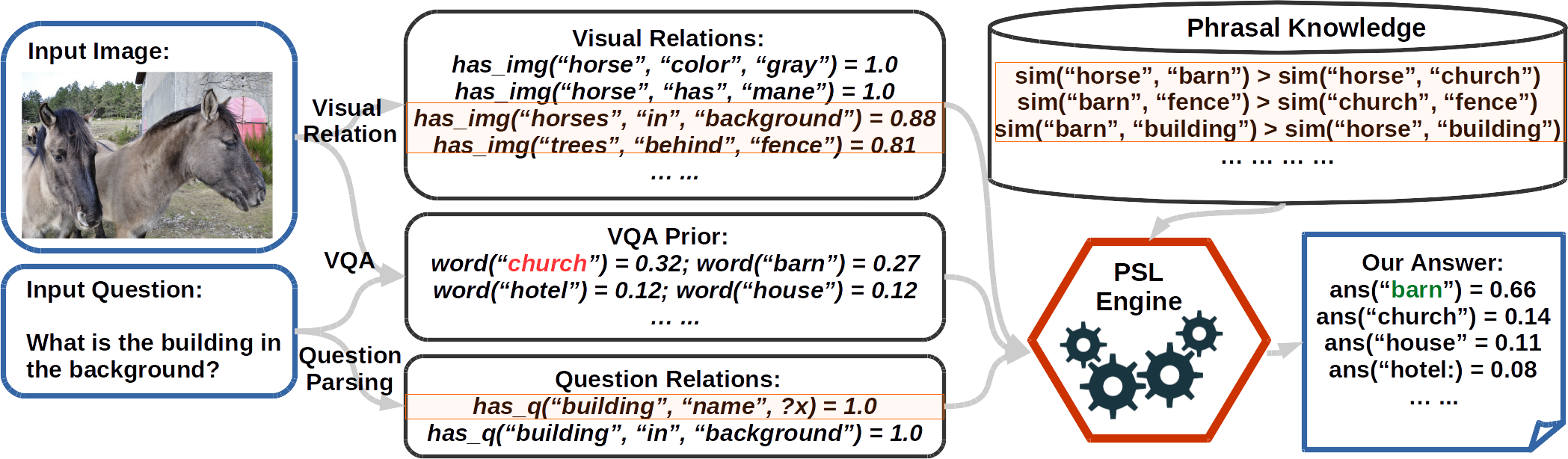} \label{fig:c4qg}}    
     \caption{An overview of the architecture followed by this paper. In this example, the reasoning engine figures out that \textit{barn} is a more likely answer, based on the evidences: i) question asks for a building and barn is a building (\textit{ontological}), ii) barn is more likely than church as it relates closely (\textit{distributional}) to other concepts in the image: \textit{horses, fence} detected from Dense Captions. Such ontological and distributional knowledge is obtained from ConceptNet and word2vec. They are encoded as similarity metrics for seamless integration with PSL.}
     \label{fig:archi}
\end{figure*}

To complement the current successful end-to-end systems, we developed two major add-on components: 1) a semantic parsing module for questions and captions, and 2) an augmented reasoning engine based on 
Probabilistic Soft Logic (PSL) \cite{bach2015hinge}. The rationale behind adding these two components are mainly threefold. Firstly, the semantic parser for question understanding helps the system to represent the information suitably for the reasoning engine; 
and the semantic parser for dense captions generated from the images \cite{densecap} adds on a structured source of semantics. Secondly, questions such as ``\textit{Is the airplane about to take off?, Is it going to rain?}'' (prospective) and ``\textit{What is common between the animal in the image and an elephant?}'' (ontological)
require various kinds of background and commonsense knowledge to answer. To reason with such knowledge together with the probabilistic nature of image understanding outputs, we develop an augmented PSL based reasoning engine. Most importantly, with the question understanding component and the reasoning engine, we are able to track the intermediate outputs (see Figure~\ref{fig:archi}) for interpreting the system itself. These intermediate outputs along with the generated evidential predicates show a promising pathway to conduct insightful performance analytics, which is incredibly difficult with existing end-to-end technologies. Thus, the presented augmentations can help the community to gain insight behind the answers, and take a step towards explainable AI \cite{DBLP:journals/corr/RibeiroSG16,lombrozo2012explanation}.

While an explicit reasoning layer is novel, there are other works that studied the reasoning aspect of VQA. Very recently, researchers have started exploring the role of language understanding and multiple-step compositional reasoning for VQA \cite{johnson2016clevr}. Instead of working on unconstrained images from original VQA corpus \cite{antol2015vqa},  the researchers switched to collecting a new corpus under a constrained setting. While the questions are designed to track aspects of multi-step reasoning, the constrained setting reduces the noise introduced by the image understanding pipelines, and simplifies the challenge that a reasoning module might face in an unconstrained environment. 
Instead, our reasoning system aims to deal with the vast amount of recognition noises introduced by image understanding systems, and targets solving the VQA task over unconstrained (natural) images. 
The presented reasoning layer is a generic engine that can be adapted to solve other image understanding tasks that require explicit reasoning. 
We intend to make the details about the engine publicly available for further research. 

Here we highlight our contributions: i) we present a novel reasoning component that successfully infers answers from various (noisy) knowledge sources for (primarily \textit{what} and \textit{which}) questions posed on unconstrained images; ii) The reasoning component is an augmentation of the PSL engine to  reason using phrasal similarities, which by its nature can be used for other language and vision tasks; iii) we annotate a subset of Visual Genome \cite{krishnavisualgenome} captions with word-pairs and open-ended relations, which can be used as the seed data for semi-supervised semantic parsing of captions. 










\section{Related Work}

Our work is influenced by four thrusts of work: i) predicting structures from images (scene graph/visual relationship), ii) predicting structures from natural language (semantic parsing), iii) QA on structured knowledge bases; and the target application area of Visual Question Answering. 

 \textbf{Visual Relationship Detection or Scene Graphs:}  Recently, several approaches have been proposed to obtain structured information from static images. \cite{elliott2013image} uses objects and spatial relations between them to represent the spatial information in images, as a graph. \cite{927} uses open-ended phrases (primarily semantic, actions, linking verbs and spatial relations) as relations between all the objects and regions (nouns) to represent the scene information as a scene graph. \cite{lu2016visual} predicts visual relationships 
 from images to represent a set of spatial and semantic relations between objects, and regions. To answer questions about an image, we need both the semantic and spatial relations between objects, regions, and their attributes (such as, $\langle\text{person, wearing, shirt}\rangle$, $\langle\text{person, standing near, pool}\rangle$, and $\langle\text{ shirt, color, red}\rangle$). Defining a closed set of meaningful relations to encode the required knowledge from perception (or language) falls under the purview of semantic parsing and is an unsolved problem. Current state-of-the-art systems use a large set of open-ended phrases as relations, and learn relationship triplets in an end-to-end manner.
 
\textbf{Semantic Parsing:} 
  Researchers in NLP have pursued various approaches to formally represent the meaning of a sentence. They can be categorized based on the (a) breadth of the application: i) general-purpose semantic parsers ii) application specific (for QA against structured Knowledge bases); and (b) the target representation, such as: i) logical languages ($\lambda$-calculus \cite{DBLP:journals/corr/Rojas15}, first order logic), and ii) structured semantic graphs.  Our processing of questions and captions is more closely related to the general-purpose  parsers that represent a sentence using a logical language or labeled graphs, also represented as a set of triplets $\langle node_1, relation, node_2\rangle$. In the first range of systems, the Boxer parser \cite{bos2008wide}, translates English sentences into first order logic. Despite its many advantages, this parser fails to represent the event-event and event-entity relations in the text. 
  Among the second category, there are many parsers which proposes to convert English sentences into the AMR representation \cite{banarescu2013abstract}. However, the available parsers are somewhat erroneous. Other semantic parsers such as K-parser \cite{DBLP:conf/ijcai/SharmaVAB15}, represent sentences using meaningful relations. But they are also error-prone.

\textbf{QA on Structured Knowledge Bases:}   Our reasoning approach is motivated by the graph-matching approach, often followed in Question-Answering systems on structured databases \cite{DBLP:conf/emnlp/BerantCFL13,DBLP:conf/kdd/FaderZE14}.  In this methodology, a question-graph is created, that has a node with a missing-label ($?x$). Candidate queries are generated based on the predicted semantic graph of the question. Using these queries (database queries for Freebase QA), candidate entities (for $?x$) are retrieved. From structured Knowledge-bases (such as Freebase), or, unstructured text, candidate semantic graphs for the corresponding candidate entities are obtained. Using a ranking metric, the correct semantic graph and the answer-node is then chosen. In \cite{Molla:2006:LGQ:1654758.1654768}, authors learn graph-based QA rules to solve factoid question answering. But, the proposed approach depends on finding maximum common sub-graph, which is highly sensitive to noisy prediction and dependent on robust closed set of nodes and edge-labels. Until recently, such top-down approaches have been difficult to attempt for QA in images. However, recent advancements of object, attributes and relationship detections has opened up the possibility of efficiently detecting structures from images and applying reasoning on these structures. 

In the field of \textbf{Visual Question Answering}, very recently, researchers have spent a significant amount of effort on creating datasets and proposing models of visual question answering \cite{antol2015vqa,malinowski2015ask,gao2015you,ma2015learning,adityadeepiu}. Both \cite{antol2015vqa} and \cite{gao2015you} adapted MS-COCO \cite{lin2014microsoft} images and created an open domain dataset with human generated questions and answers. To answer questions about images both \cite{malinowski2015ask} and \cite{gao2015you} use recurrent networks to encode the sentence and output the answer. Specifically, \cite{malinowski2015ask} applies a single network to handle both encoding and decoding, while \cite{gao2015you} divides the task into an encoder network and a decoder one. More recently, the work from \cite{ren2015image} formulates the task straightforwardly as a classification problem and focuses on the 
questions that can be answered with one word.
 
A survey article \cite{wu2016visual} on VQA dissects the different methods into the following categories: i) Joint Embedding methods, ii) Attention Mechanisms, iii) Compositional Models, and iv) Models using External Knowledge Bases. Joint Embedding approaches were first used in image captioning methods where the text and images are jointly embedded in the same vector space. For VQA, primarily a Convolutional Neural Network for images and a Recurrent Neural Network for text is used to embed into the same space and this combined representation is used to learn the mapping between the answers and the question-and-images space. Approaches such as \cite{malinowski2015ask,gao2015you} fall under this category. \cite{zhu2015visual7w,lu2016hierarchical,andreas2015deep} use different types of attention mechanisms (word-guided, question-guided attention map etc) to solve VQA. Compositional Models take a different route and try to build reusable smaller modules that can be put together to solve VQA. Some of the works along this line are Neural Module Networks (\cite{andreas2015deep}), and Dynamic Memory Networks (\cite{DBLP:journals/corr/KumarISBEPOGS15}). Lately, there have been attempts of creating QA datasets that solely comprises of questions that require additional background knowledge along with information from images \cite{wang2015explicit}.

   In this work, to answer a question about an image, we add a probabilistic reasoning mechanism on top of the knowledge (represented as semantic graphs) extracted from the image and the question. To extract such graphs, 
   we use semantic parsing on generated dense captions from the image, and the natural language question. To minimize the error in parsing, we use a large set of open-ended phrases as relations, and simple heuristic rules to predict such relations. To resolve the semantics of these open-ended arguments, we use knowledge about words (and phrases) in the probabilistic reasoning engine. In the following section, we introduce the knowledge sources and the reasoning mechanism used.
   
\section{Knowledge and Reasoning Mechanism}
   In this Section, we briefly introduce the additional knowledge sources used for reasoning on the semantic graphs from question and the image; and the reasoning mechanism used to reason about the knowledge. As we use open-ended phrases as relations and nodes, we need knowledge about phrasal similarities. We obtain such knowledge from the learnt word-vectors using word2vec.
   
   {\bf Word2vec} uses distributional semantics to capture word meanings and produces fixed-length word embeddings (vectors). These pre-trained word-vectors  have been successfully used in numerous NLP applications and the induced vector-space is known to capture the graded similarities between words with reasonable accuracy \cite{mikolov2013efficient}. In this work, we use the $3$ Million word-vectors trained on Google-News corpus \cite{mikolov2013efficient}.
   
    To reason with such knowledge we explored various reasoning formalisms and found Probabilistic Soft Logic (PSL)
\cite{bach2015hinge} to be the most suitable, as it can not only handle relational structure, inconsistencies and uncertainty, thus allowing one to express rich probabilistic graphical models (such as  Hinge-loss Markov random fields), but it also seems to scale up better than its alternatives such as Markov Logic Networks \cite{Richardson:2006:MLN:1113907.1113910}.
   
\subsection{Probabilistic Soft Logic (PSL)}   
A PSL model is defined using a set of weighted if-then rules in first-order logic.  
For example, from \cite{bach2015hinge} we have:
\begin{equation*}
\small
\begin{aligned}
0.3: votesFor(X,Z) &\leftarrow ~friend(X,Y) \land votesFor(Y,Z)\\
0.8: votesFor(X,Z) &\leftarrow spouse(X,Y) \land votesFor(Y,Z) 
\end{aligned}
\end{equation*}

 In this notation, we use upper case letters to represent variables and lower case letters for constants. The above rules applies to all $X,Y,Z$, for which the predicates have non-zero truth values. The weighted rules encode the knowledge that a person is more likely to vote for the same person as his/her spouse than the person that his/her friend votes for. 
  In general,  let $\bm{C}=(C_1,...,C_m)$ be such a collection of weighted rules where each $C_j$ is a disjunction of literals, where each literal is a variable $y_i$ or its negation $\neg y_i$, where $y_i\in \bm{y}$. Let $I_j^{+}$ (resp. $I_j^{-}$ ) be
  the set of indices of the variables that are not negated (resp. negated) in $C_j$. Each $C_j$ can be represented as: 
\begin{equation}  
w_j : \lor_{i \in I_j^{+}} y_i \leftarrow \land_{i \in I_j^{-}} y_i ,
\end{equation}

or equivalently, $w_j: \lor_{i \in I_j^{-}} (\neg y_i) \bigvee \lor_{i \in I_j^{+}} y_i$. 
  A rule $C_j$ is associated with a non-negative weight $w_j$. PSL relaxes the boolean truth values of each ground atom $a$ (constant term or predicate with all variables replaced by constants) to the interval [0, 1], denoted as $V(a)$. To compute soft truth
values, 
Lukasiewicz's relaxation \cite{klir1995fuzzy} of conjunctions ($\land$), disjunctions ($\lor$) and negations ($\neg$) 
are used:
\begin{equation*}
\small
\begin{aligned}
V(l_1\land l_2) = max\{0,V(l_1)+V(l_2)-1\}\\
V(l_1\lor l_2) = min\{1,V(l_1)+V(l_2)\}\\
V(\neg l_1) = 1-V(l_1).
\end{aligned}
\end{equation*}

In PSL, the ground atoms are considered as random variables, and the joint distribution is modeled using Hinge-Loss Markov Random Field (HL-MRF). An HL-MRF is defined as follows:
 Let $\bm{y}$ and $\bm{x}$ be two vectors of $n$ and ${n'}$ random variables respectively, over the domain $D =[0,1]^{n+{n'}}$. The feasible set $\tilde{D}$ is a subset of $D$, which satisfies a set of inequality constraints over the random variables.

A \textit{Hinge-Loss Markov Random Field} $\mathbb{P}$ is a probability density over $D$, defined as: if $(\bm{y},\bm{x}) \notin \tilde{D}$, then $\mathbb{P}(\bm{y}|\bm{x})=0$; if $(\bm{y},\bm{x}) \in \tilde{D}$, then:
\begin{equation}
\begin{aligned}
\mathbb{P}(\bm{y}|\bm{x}) \propto  exp(-f_{\bm{w}}(\bm{y},\bm{x})).
\end{aligned}
\end{equation}

  In PSL, the hinge-loss energy function $f_{\bm{w}}$ is defined as:
\begin{equation*}
f_{\bm{w}}(\bm{y}) = \sum\limits_{C_j \in \bm{C}} w_j\text{ }max\big\{ 1- \sum_{i \in I_j^{+}} V(y_i) - \sum_{i \in I_j^{-}} (1- V(y_i)),0\big\}.
\end{equation*} 
The maximum-a posteriori (MAP) inference objective of PSL becomes:
{\small
\begin{equation}
\begin{aligned}
\setlength{\abovedisplayskip}{0pt}
\mathbb{P}(\bm{y}) &\equiv \argmax_{\bm{y}\in [0,1]^n} exp(-f_{\bm{w}}(\bm{y})) \\
 &\equiv \argmin_{\bm{y}\in [0,1]^n} \sum\limits_{C_j \in \bm{C}} w_j\text{ }max\Big\{ 1- \sum_{i \in I_j^{+}} V(y_i) \\
&- \sum_{i \in I_j^{-}} (1-V(y_i)),0\Big\},
\label{eq:4}
\end{aligned}
\end{equation}
}
  where the term $w_j\times max\big\{ 1- \sum_{i \in I_j^{+}} V(y_i) - \sum_{i \in I_j^{-}} (1- V(y_i)),0\big\}$ measures the ``distance to satisfaction'' for each rule $C_j$.

\section{Our Approach}
    Inspired by the textual Question-Answering systems \cite{DBLP:conf/emnlp/BerantCFL13,Molla:2006:LGQ:1654758.1654768}, we adopt the following approach: i) we first detect and extract relations between objects, regions and attributes (represented using $has\_img(w_1,rel,w_2)$\footnote{In case of images, $w_1$ and $w_2$ belong to the set of objects, regions and attributes seen in the image. In case of questions, $w_1$ and $w_2$ belong to the set of nouns and adjectives. For both, $rel$ belongs to set of open-ended semantic, spatial relations, obtained from the Visual Genome dataset.}) from images, constituting $G_{img}$; ii) we then extract relation between nouns, the Wh-word and adjectives (represented using $has\_q(w_1,rel,w_2)$) from the question (constituting $G_q$), where the relations in both come from a large set of open-ended relations; and iii) we reason over the structures using an augmented reasoning engine that we developed. Here,  we use PSL, as it is well-equipped to reason with soft-truth values of predicates and it scales well \cite{bach2015hinge}.

\subsection{Extracting Relationships from Images}
   We represent the factual information content in images using relationship triplets\footnote{Triplets are often used to represent knowledge, such as RDF-triplets (in semantic web), triplets in Ontological knowledge bases has the form $\langle subject, predicate, object\rangle$ \cite{wang2016fvqa}. Triplets in \cite{lu2016visual} use $\langle object_1, predicate, object_2\rangle$ to represent visual information in images.}. To answer factual questions such as ``what color shirt is the man wearing'', ``what type of car is parked near the man'', we need relations such as \textit{color, wearing, parked near}, and \textit{type of}. In summary, to represent the factual information content in images as triplets, we need semantic relations, spatial relations, and action and linking verbs between objects, regions and attributes (i.e. nouns and adjectives). 
   
   To generate relationships from an image, we use the pre-trained Dense Captioning system \cite{densecap} to generate dense captions (sentences) from an image, and heuristic rule-based semantic parsing module to obtain relationship triplets. For semantic parsing, we detect nouns and noun phrases using a syntactic parser (we use Stanford Dependency parsing \cite{de2006generating}). For target relations, we use a filtered subset\footnote{We removed noisy relations with spelling mistakes, repetitions, and noun-phrase relations.} of open-ended relations from the Visual Genome dataset \cite{krishnavisualgenome}. To detect the relations between two objects or, object and an attribute (nouns, adjectives), we extract the connecting phrase from the sentence and the connecting nodes in the shortest dependency path from the dependency graph\footnote{The shortest path hypothesis \cite{Xu2016EnhancingFQ} has been used to detect relations between two nominals in a sentence in textual QA. Primarily, the nodes in the path and the connecting phrase construct semantic and syntactic feature for the supervised classification. However, as we do not have a large annotated training data and the set of target relations is quite large ($20000$), we resort to heuristic phrase similarity measures. These measures work better than a semi-supervised iterative approach.}. We use word-vector based phrase similarity (aggregate word-vectors and apply cosine similarity) to detect the most similar phrase as a relation. 
   To verify this heuristic approach, we manually annotated $4500$ samples using the region-specific captions provided in the Visual Genome dataset. The heuristic rule-base approach achieves a $64\%$ exact-match  accuracy over $20102$ possible relations. We provide some example annotations and predicted relations in Table \ref{tab:parsing}. 
 
\begin{table}[!htb]
\centering
\resizebox{\columnwidth}{!}{%
\begin{tabular}{llll}
\hline
\multicolumn{1}{l|}{Sentence}                                                                                                  & \multicolumn{1}{l|}{Words}                   & \multicolumn{1}{l|}{Annotated}      & \multicolumn{1}{l}{Predicted}      \\ \hline
\multicolumn{1}{l|}{\multirow{2}{*}{\begin{tabular}[c]{@{}l@{}}cars are parked on \\ the side of the road\end{tabular}}}       & \multicolumn{1}{l|}{{[}'cars', 'side'{]}}    & \multicolumn{1}{l|}{parked on the}  & \multicolumn{1}{l}{parked on}      \\ \cline{2-4} 
\multicolumn{1}{l|}{}                                                                                                          & \multicolumn{1}{l|}{{[}'cars', 'road'{]}}    & \multicolumn{1}{l|}{parked on side} & \multicolumn{1}{l}{on its side in} \\ \hline
\multicolumn{1}{l|}{\begin{tabular}[c]{@{}l@{}}there are two men \\ conversing in the photo\end{tabular}}                      & \multicolumn{1}{l|}{{[}'men', 'photo'{]}}    & \multicolumn{1}{l|}{in}             & \multicolumn{1}{l}{conversing in}  \\ \hline
\multicolumn{1}{l|}{\begin{tabular}[c]{@{}l@{}}the men are on \\ the sidewalk\end{tabular}}                                    & \multicolumn{1}{l|}{{[}'men', 'sidewalk'{]}} & \multicolumn{1}{l|}{on}             & \multicolumn{1}{l}{on}             \\ \hline
\multicolumn{1}{l|}{\begin{tabular}[c]{@{}l@{}}the trees do not \\ have leaves\end{tabular}}                                   & \multicolumn{1}{l|}{{[}'trees', 'leaves'{]}} & \multicolumn{1}{l|}{do not have}    & \multicolumn{1}{l}{do not have}    \\ \hline
\multicolumn{1}{l|}{\begin{tabular}[c]{@{}l@{}}there is a big clock \\ on the pole\end{tabular}}                               & \multicolumn{1}{l|}{{[}'clock', 'pole'{]}}   & \multicolumn{1}{l|}{on}             & \multicolumn{1}{l}{on}             \\ \hline
\multicolumn{1}{l|}{\multirow{2}{*}{\begin{tabular}[c]{@{}l@{}}a man dressed in \\ a red shirt and black pants.\end{tabular}}} & \multicolumn{1}{l|}{{[}'man', 'shirt'{]}}    & \multicolumn{1}{l|}{dressed in}     & \multicolumn{1}{l}{dressed in}     \\ \cline{2-4} 
\multicolumn{1}{l|}{} & \multicolumn{1}{l|}{{[}'man', 'pants'{]}}    & \multicolumn{1}{l|}{dressed in}     & \multicolumn{1}{l}{dressed in}     \\ \hline                              
\end{tabular}
}
\caption{Example Captions, Groundtruth Annotations and Predicted Relations between words.}
\label{tab:parsing}
\end{table}

\begin{table}[!htb]
   \centering
    \resizebox{\columnwidth}{!}{%
\begin{tabular}{|l|l|l|}
\hline
\{\bf Predicates\}          & \{\bf Semantics\}                                                                                    & \{\bf Truth Value\}                                                                      \\ \hline\hline
$word(Z)$                   & Prior of Answer $Z$                                                                                 & 1.0 or VQA prior                                                                         \\ \hline
$has\_q(X,R,Y)$             & Triplet from the Question                                                                            & From Relation Prediction                                                                 \\ \hline
$has\_img(X1,R1,Y1)$        & Triplet from Captions                                                                                & \begin{tabular}[c]{@{}l@{}}From Relation Prediction \\ and Dense Captioning\end{tabular} \\ \hline
\begin{tabular}[c]{@{}l@{}}$has\_img\_ans(Z,$\\ $X1,R1,Y1)$ \end{tabular} & \begin{tabular}[c]{@{}l@{}}Potential involving the answer $Z$ \\ with respect to image triplet\end{tabular} & Inferred using PSL                                                                       \\ \hline
$candidate(Z)$              & Candidate Answer  $Z$                                                                                   & Inferred using PSL                                                                       \\ \hline
$ans(Z)$                    & Final Answer  $Z$                                                                                       & Inferred using PSL                                                                       \\ \hline
\end{tabular}
}
    \caption{List of predicates involved and the sources of the soft truth values.}
    \label{input_vqa_psl}
\end{table}

\subsection{Question Parsing}
   For parsing questions, we again use the Stanford Dependency parser to extract the nodes (nouns, adjectives and the Wh question word). For each pair of nodes, we again extract the linking phrase and the shortest dependency path; and, use phrase-similarity measures to predict the relation. The phrase-similarity is computed as above. After this phase, we construct the input predicates for our rule-based Probabilistic Soft Logic engine.

\subsection{Logical Reasoning Engine}

   Finally based on the set of triplets, we use a probabilistic logical reasoning module. 
   
      Given an image $I$ and a question $Q$, we rank the candidate answers $Z$ by estimating the conditional probability of the answer, i.e. $P(Z|I,Q)$. In PSL, to formulate such a conditional probability function, we use the (non-negative) truth values of the candidate answers and pose an upper bound on the sum of the values over all answers. Such a constraint can be formulated based on the PSL optimization formulation.

   \textbf{PSL: Adding the Summation Constraint}:  
   As described earlier, for a database $\bm{C}$ consisting of the rules $C_j$, 
   the underlying optimization formulation for the inference problem is given in Equation \ref{eq:4}.
   In this formulation, $\bm{y}$ is the collection of observed and unobserved ($\bm{x}$) variables. A summation constraint over the unobserved variables ($\sum_{x \in \bm{x}} V(x) \leq S$) forces the optimizer to find a solution, where the most probable variables are assigned higher truth values.

\begin{equation}
\sum_{y \in \bm{y}} V(y) \leq S
\end{equation}
   
   \textbf{Input}:  The triplets from the image and question constitute $has\_img()$ and $has\_q()$ tuples. For $has\_img()$, the confidence score is computed using the confidence of the dense caption and the confidence of the predicted relation. For $has\_q()$, only the  similarity of the predicted relation is considered. We also input the set of answers as $word()$ tuples. The truth values of these predicates define the prior confidence of these answers. It can come from weak to strong sources (frequency, existing VQA system etc.). The list of inputs is summarized in Table \ref{input_vqa_psl}.  

  \textbf{Formulation}: Ideally, the sub-graphs related to the answer-candidates can be compared directly to the semantic graph of the question and the corresponding missing information ($?x$) can then be found. However, due to noisy detections and the inherent complexities (such as paraphrasing) in natural language, such a strong match is not feasible. We relax this constraint by using the concept of ``soft-firing''\footnote{If $a \land b \land c \implies d$ with some weight, then with some weight $a \implies d$.}  and incorporating knowledge of phrase-similarity in the reasoning engine.
     
      As the answers ($Z$) are not guaranteed to be present in the captions, we calculate the \textit{relatedness} of each image-triplet ($\langle X,R1,Y1 \rangle$) to the answer, modeling the potential $\phi(Z,\langle X,R1,Y1 \rangle)$. Together, with all the image-triplets, they model the potential involving $Z$ and $G_{img}$.  
      For ease of reading, we use $\approx_{p}$ notation to denote the phrase similarity function.
 \begin{equation*}
 \small
\begin{aligned}
w_1: &~has\_img\_ans(Z,X,R1,Y1) \leftarrow \\
&\land word(Z) \land has\_img(X,R1,Y1) \\ 
&\land Z \approx_{p} X \land Z \approx_{p} Y1.
\end{aligned}
 \end{equation*} 
  We then add rules to predict the candidate answers ($candidate(.)$) by using fuzzy matches with image triplets and the question triplets;  they model the potential involving $Z,G_{img}$ and $G_q$ collectively. 
 \begin{equation*}
 \small
\begin{aligned}
w_2: &~candidate(Z) \leftarrow word(Z).\\
w_3: &~candidate(Z) \leftarrow word(Z) \\
&\land has\_q(Y,R,X) \\
&\land has\_img\_ans(Z,X1,R1,Y1) \\
&\land R \approx_{p} R1 \land Y \approx_{p} Y1 \land X \approx_{p} X1 .
\end{aligned}
 \end{equation*}
  Lastly, we match the question-triplet with missing node-labels.
  \begin{equation*}
  \small
\begin{aligned}
w_4: &~ans(Z) \leftarrow has\_q(X,R,?x) \\
&\land  has\_img(Z,R,X) \land candidate(Z).\\
w_5: &~ans(Z) \leftarrow has\_q(X,R,?x) \\
&\land  has\_img(Z1,R,X) \land candidate(Z)\\
& \land Z \approx_{p} Z1. \\
w_6: &~ans(Z) \leftarrow has\_q(X,R,?x) \\
&\land  has\_img(Z1,R1,X1) \land candidate(Z)\\
& \land Z \approx_{p} Z1 \land R \approx_{p} R1 \land X \approx_{p} X1.
\end{aligned}
 \end{equation*} 
    We use a summation constraint over $ans(Z)$ to force the optimizer to increase the truth value of the answers which satisfies the most rules. Our system learns the rules' weights using the Maximum Likelihood method \cite{bach2015hinge}. 
   
\section{Experiments}

To validate that the presented reasoning component is able to improve existing image understanding systems and do better robust question answering with respect to unconstrained images, we adopt the standard VQA dataset to serve as the test bed for our systems. In the following sections, we start from describing the benchmark dataset, followed by two experiments we conducted on the dataset. We then discuss the experimental results and state why they validate our claims.

\subsection{Benchmark Dataset}

MSCOCO-VQA \cite{antol2015vqa} is the largest VQA dataset that contains both multiple choices and open-ended questions about arbitrary images collected from the Internet. This dataset contains $369,861$ questions and $3,698,610$ ground truth answers based on $123,287$ MSCOCO images. These questions and answers are sentence-based and open-ended. The training and testing split follows MSCOCO-VQA official split. Specifically, we use $82,783$ images for training and $40,504$ validation images for testing. We use the validation set to report question category-wise performances for further analysis. 


\begin{table}[!htb]
\centering
\resizebox{\columnwidth}{!}{%
\begin{tabular}{l|l|l|l|l}
\hline
\multicolumn{1}{l|}{}                                                                     & Categories                     & CoAttn & PSLDVQ           & \begin{tabular}[c]{@{}l@{}}PSLDVQ-\\ +CN\end{tabular}       \\ \hline
\multirow{10}{*}{\begin{tabular}[c]{@{}l@{}}Speci-\\ fic\end{tabular}}                     & what animal is (516)           & 65     & \textbf{66.22} & \textbf{66.36} \\
                                                                                           & what brand  (526)              & 38.14  & 37.51            & 37.55            \\
                                                                                           & what is the man (1493)         & 54.82  & \textbf{55.01} & 54.66            \\
                                                                                           & what is the name  (433)        & 8.57   & 8.2              & 7.74            \\
                                                                                           & what is the person (500)       & 54.84  & \textbf{54.98} & 54.2            \\
                                                                                           & what is the woman (497)        & 45.84  & \textbf{46.52} & 45.41            \\
                                                                                           & what number is  (375)          & 4.05   & \textbf{4.51}  & \textbf{4.67}  \\
                                                                                           & what room is (472)             & 88.07  & 87.86            & \textbf{88.28}\\
                                                                                           & what sport is  (665)           & 89.1   & \textbf{89.1}  & 89.04            \\
                                                                                           & what time (1006)               & 22.55  & 22.24            & 22.54            \\ \hline
\multicolumn{1}{l|}{\multirow{3}{*}{\begin{tabular}[c]{@{}l@{}}Sum-\\ mary\end{tabular}}} & \textbf{Other}               & 57.49  & \textbf{57.59} & 57.37            \\  
\multicolumn{1}{l|}{}                                                                     & \textbf{Number}              & 2.51   & \textbf{2.58}  & \textbf{2.7}   \\  \cline{2-5}
\multicolumn{1}{l|}{}                                                                     & \textbf{Total}               & 48.49  & \textbf{48.58} & 48.42            \\ \hline \hline
\multirow{5}{*}{\begin{tabular}[c]{@{}l@{}}Color\\ Related\end{tabular}}                   & what color (791)               & 48.14  & 47.51            & 47.07            \\
                                                                                           & what color are the (1806)      & 56.2   & 55.07            & 54.38            \\
                                                                                           & what color is (711)            & 61.01  & 58.33            & 57.37            \\
                                                                                           & what color is the (8193)       & 62.44  & 61.39            & 60.37            \\
                                                                                           & what is the color of the (467) & 70.92  & 67.39            & 64.03            \\ \hline 
\multirow{17}{*}{\begin{tabular}[c]{@{}l@{}}Gener-\\ al\end{tabular}}                      & what (9123)                    & 39.49  & 39.12            & 38.97            \\
                                                                                           & what are (857)                 & 51.65  & \textbf{52.71} & \textbf{52.71} \\
                                                                                           & what are the  (1859)           & 40.92  & 40.52            & 40.49            \\
                                                                                           & what does the (1133)           & 21.87  & 21.51            & 21.49            \\
                                                                                           & what is (3605)                 & 32.88  & \textbf{33.08} & 32.65            \\
                                                                                           & what is in the (981)           & 41.54  & 40.8             & 40.49            \\
                                                                                           & what is on the (1213)          & 36.94  & 35.72            & 35.8             \\
                                                                                           & what is the (6455)             & 41.68  & 41.22            & 41.4            \\
                                                                                           & what is this (928)             & 57.18  & 56.4             & 56.25            \\
                                                                                           & what kind of      (3301)       & 49.85  & 49.81            & 49.84            \\
                                                                                           & what type of (2259)            & 48.68  & 48.53            & \textbf{48.77} \\
                                                                                           & where are the (788)            & 31     & 29.94            & 29.06            \\
                                                                                           & where is the (2263)            & 28.4   & 28.09            & 27.69            \\
                                                                                           & which (1421)                   & 40.91  & \textbf{41.2}  & 40.73            \\
                                                                                           & who is (640)                   & 27.16  & 24.11            & 21.91           \\
                                                                                           & why (930)                      & 16.78  & 16.54            & 16.08            \\
                                                                                           & why is the (347)               & 16.65  & 16.53            & \textbf{16.74} \\ \hline 
\end{tabular}
}
\caption{Comparative 
results on the VQA validation questions. We report results on the non-Yes/No and non-Counting question types. Highest accuracies achieved by our system is presented in bold. We report the summary results of the set of ``specific'' question categories.}
\label{tab:results}
\end{table}

\begin{figure*}[htb!]
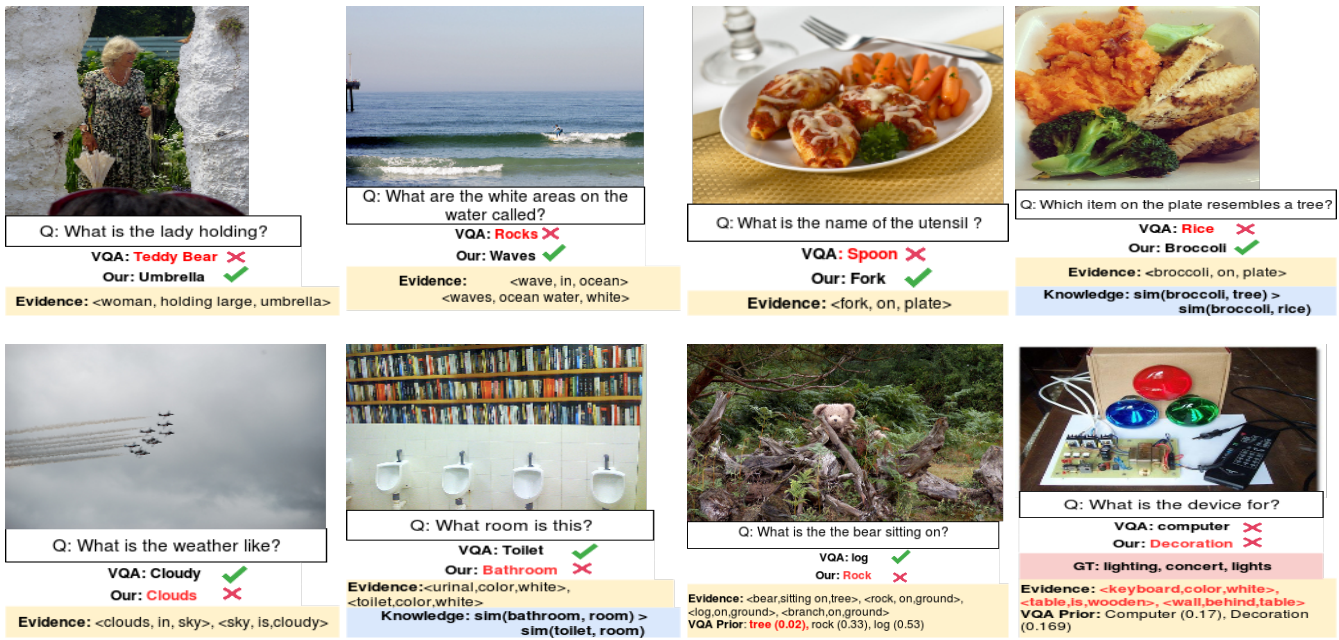

     \centering      \subfloat{\includegraphics[width=0.25\textwidth,height=0.18\textheight]{figures_vqa/umbrella_pos.png}}
      \hfill      \subfloat{\includegraphics[width=0.25\textwidth,height=0.18\textheight]{figures_vqa/waves_pos.png}}      
       \hfill      \subfloat{\includegraphics[width=0.24\textwidth,height=0.18\textheight]{figures_vqa/fork1_pos.png}}
      \hfill
\subfloat{\includegraphics[width=0.24\textwidth,height=0.18\textheight]{figures_vqa/broccoli_pos.png} }
\hfill
 \subfloat{\includegraphics[width=0.25\textwidth,height=0.17\textheight]{figures_vqa/cloudy_neg.png}}
      \hfill      \subfloat{\includegraphics[width=0.25\textwidth,height=0.17\textheight]{figures_vqa/toilet_neg.png}}      
       \hfill      \subfloat{\includegraphics[width=0.24\textwidth,height=0.17\textheight]{figures_vqa/bear_neg.png}}
      \hfill
\subfloat{\includegraphics[width=0.24\textwidth,height=0.17\textheight]{figures_vqa/traffic_light_neg.png} }
     \caption{Positive and Negative results generated by our reasoning engine. For evidence, we provide predicates that are key evidences to the predicted answer. \textbf{*}Interestingly in the last example, all 10 ground-truth answers are different. Complete end-to-end examples can be found in \url{visionandreasoning.wordpress.com}.}
     \label{fig:example_vqa}
\end{figure*}

\subsection{Experiment I: End-to-end Accuracy}
   In this experiment, we test the end-to-end accuracy of the presented PSL-based reasoning system. We use several variations as follows:
   
\noindent $\bullet$ \textbf{PSLD(ense)VQ}: Uses captions from Dense Captioning \cite{densecap} and prior probabilities from a trained VQA system  \cite{lu2016hierarchical} as truth values of answer $Z$ ($word(Z)$).
   
\noindent   $\bullet$ \textbf{PSLD(ense)VQ+CN}: We enhance PSLDenseVQ with the following. In addition to word2vec embeddings, we use the embeddings from ConceptNet 5.5  \cite{havasi2007conceptnet} to compute phrase similarities  ($\approx_p$), using the aggregate word vectors and cosine similarity. Final similarity is the average of the two similarities from word2vec and ConceptNet.
   
\noindent   $\bullet$ \textbf{CoAttn}:  We use the  output from the  hierarchical co-attention system trained by \citeauthor{lu2016visual} 2016, as the baseline system to compare.

   We use the 
   evaluation script by \cite{antol2015vqa} to evaluate accuracy on the validation data. The comparative results for each question category is presented in Table \ref{tab:results}.
   
   \textbf{Choice of question Categories}: Different question categories often require different form of background knowledge and reasoning mechanism. For example, ``Yes/No'' questions are equivalent to entailment problems (verify a statement based on information from image and background knowledge), and ``Counting'' questions are mainly recognition questions (requiring limited reasoning only to understand the question). In this work, we use semantic-graph matching based reasoning process that is often targeted to find the missing information (the label $?x$) in the semantic graph. Essentially, with this reasoning engine, we target \textit{what} and \textit{which} questions, to validate how additional structured information from captions and background knowledge can improve VQA performance. In Table \ref{tab:results}, we report and further group all the non-Yes/No and non-Counting questions into \textit{general, specific} and \textit{color} questions. We observe from Table \ref{tab:results} that the majority of the performance boost is with respect to the questions targeting specific types of answers. When dealing with other general or color related questions, adding the explicit reasoning layer helps in limited number of questions. \textit{Color} questions are recognition-intensive questions. In cases where the correct color is not detected, reasoning can not improve performance. For \textit{general} questions, the rule-base requires further exploration. 
   For \textit{why} questions, often there could be multiple answers, prone to large linguistic variations. Hence the evaluation metric requires further exploration.

\subsection{Experiment II: Explicit Reasoning}

  In this experiment, we discuss the examples where explicit reasoning helps predict the correct answer even when detections from the end-to-end VQA system are noisy. We provide these examples in Figure \ref{fig:example_vqa}. As shown, the improvement comes from the additional information from captions, and usage of background knowledge. We provide key evidence predicates that helps the reasoning engine to predict the correct answer. However, the quantitative evaluation of such evidences is still an open problem. 
  Nevetheless, one primary advantage of our system is its ability to generate the influential key evidences that lead to the final answer, and being able to list them as (structured) predicates\footnote{We can simply obtain the predicates  in the body of the grounded rules that were satisfied (i.e. distance to satisfaction is zero) by the inferred predicates.}.  The examples in Figure \ref{fig:example_vqa} includes key evidence predicates and knowledge predicates used. We will make our final answers together with ranked key evidence predicates publicly available for further research.

\subsection{Experiment III: An Adversarial Example}
\begin{figure}[!h]
     \centering
 \includegraphics[width=0.47\textwidth]{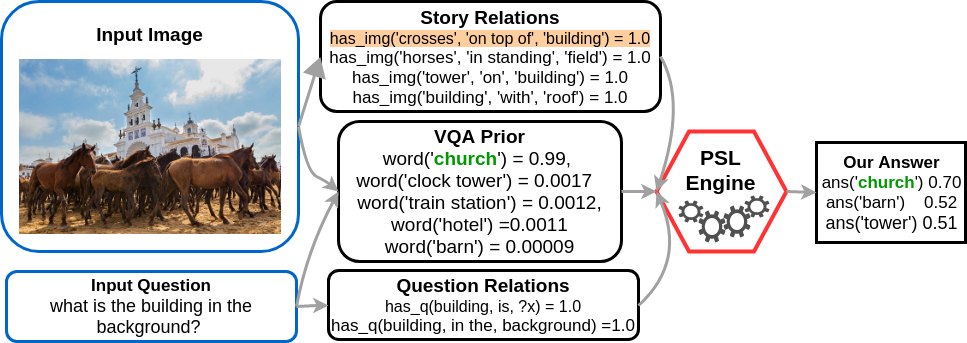}
     \caption{An adversarial example as opposed to the motivating example at Figure \ref{fig:c4qg}. The supporting predicate found is: \textit{has\_img(crosses, on top of, building)}. }
     \label{fig:adv}
\end{figure}

  Apart from understanding the natural language question, commonsense knowledge can help rectify final outcomes in essentially two situations: i) in case of noisy detections (a weak perception module) and ii) in case of incomplete information (such as occlusions). In Figure \ref{fig:c4qg}, we show a motivating example of partial occlusion, where the data-driven neural network-based VQA system predicts the answer \textit{church}, and the PSL-based reasoning engine chooses a more logical answer \textit{barn} based on cues (such as \textit{horses in the foreground}) from other knowledge sources (\textit{dense captions}). A question remains, whether the reasoning engine itself injects a bias from commonsense, i.e. whether it will predict \textit{barn}, even if there is actually a church in the background and while the commonsense knowledge still dictates that the \textit{building around the horses could be a barn}. To answer this question, we further validate our system with an adversarial example (see Figure~\ref{fig:adv}). As expected, our PSL engine still predicts the correct answer, and improves the probabilities of more probable answers (barn, tower). In addition, it also provides the evidential predicates to support the answer.
 
\section{Conclusion and Future Work}

In this paper, we present an integrated system that adopts an explicit reasoning layer over the end-to-end neural architectures. Experimental results on the visual question answering testing bed validates that the presented system is better suited for answering ``what'' and ``which'' questions where additional structured information and background knowledge are needed. We also show that with the explicit reasoning layer, our system can generate both final answers to the visual questions as well as the top ranked key evidences supporting these answers. They can serve as explanations and validate that the add-on reasoning layer improves system's overall interpretability. Overall our system achieves a performance boost over several VQA categories at the same time with an improved explainability.

Future work includes adopting different learning mechanisms to learn the weights of the rules, and the structured information from the image. We also plan to extend Inductive Logic Programming algorithms (such as XHAIL \cite{ray2003hybrid}) to learn rules for probabilistic logical languages, and scale them for large number of predicates.

\bibliography{egbib,references_somak}
\bibliographystyle{aaai}
\end{document}